\documentclass[a4paper,fleqn]{cas-sc}

\usepackage{xeCJK}

\usepackage[authoryear]{natbib}

\usepackage{xcolor}
\usepackage{amssymb}
\usepackage{amsmath}
\usepackage{makecell}

\usepackage{booktabs}
\usepackage{multicol}
\usepackage{multirow}
\usepackage{subcaption}
\usepackage{tabularx}
\usepackage{tcolorbox}
\usepackage{array} 
\usepackage{longtable}
\usepackage{geometry} 
\usepackage{placeins}

\usepackage{booktabs,siunitx, xcolor, colortbl}

\usepackage{pgfplots}
\pgfplotsset{compat=1.18}
\usepackage{placeins}

\usepackage{fontspec}

\newtcolorbox[
  auto counter,
  number within=section
]{promptbox}[2][]{
  colback=white,
  colframe=black,
  fonttitle=\bfseries,
  #1
}

\def\tsc#1{\csdef{#1}{\textsc{\lowercase{#1}}\xspace}}
\tsc{WGM}
\tsc{QE}

\begin{document}
\let\WriteBookmarks\relax
\def\floatpagepagefraction{1}
\def\textpagefraction{.001}

\shorttitle{}    

\shortauthors{}  


\title [mode = title]{Evaluating Chinese Ambiguity Understanding in Large Language Models}

\affiliation[utokyo]{organization={Graduate School of Information Science and Technology, The University of Tokyo},
            city={Tokyo},
            country={Japan}}

\affiliation[scut]{organization={School of Software Engineering, South China University of Technology},
            addressline={Guangzhou Higher Education Mega Centre, Panyu District},  
            city={Guangzhou},
            postcode={510006}, 
            state={Guangdong},
            country={China}}

\author[utokyo]{Junwen Mo} 
\ead{mo@nlab.ci.i.u-tokyo.ac.jp}
\credit{Conceptualization, Methodology, Software, Data curation, Formal analysis, Writing -- original draft, Visualization, Validation}

\author[scut]{Yuanzhi Lu}
\credit{Methodology, Software, Data curation, Formal analysis, Writing -- original draft, Visualization, Validation}
\ead{seyzluk@mail.scut.edu.cn}

\author[scut]{Yifang Xue}
\credit{Methodology, Software, Data curation, Formal analysis, Writing -- original draft, Visualization, Validation}
\ead{202230482506@mail.scut.edu.cn}

\author[2]{Ke Xu}[orcid=0000-0002-2265-756X]
\credit{Conceptualization, Methodology, Formal analysis, Data curation, Supervision, Writing -- review and editing, Project administration, Funding acquisition, Resources}

\cormark[2]
\cortext[2]{Corresponding author. E-mail: kexu@scut.edu.cn}



\author[utokyo]{Hideki Nakayama}
\ead{nakayama@ci.i.u-tokyo.ac.jp}
\credit{Supervision, Funding acquisition, Writing -- review and editing, Resources}

\begin{abstract}
Linguistic ambiguity is critical to the robustness of Large Language Models (LLMs), yet existing research focuses mostly on English, with limited attention devoted to  Chinese. Existing Chinese ambiguity datasets (e.g., CHAmbi) suffer from poor scalability. Guided by Potential Ambiguity (PA) Theory, we design a semi-automatic pipeline to construct CHA-Gen. It is the first PA Theory-grounded Chinese ambiguity dataset, which comprises 5,712 sentences (2,414 ambiguous, 3,298 unambiguous) across 18 potential ambiguous structures. Evaluating LLMs (e.g. Gemma 3, Qwen 2.5/3 series) via direct querying and machine translation, we find that LLMs struggle with ambiguity detection (improved by CoT prompting). Analysis of Qwen3-32B's CoT rationales reveals three common failure modes: ambiguity blindness, misattribution, and premature resolution. Uncertainty quantification with semantic entropy metric shows higher uncertainty for ambiguous sentences. Moreover, instruction tuning induces overconfidence, whereas Base models better capture semantic diversity. We further observe that models exhibit a bias toward dominant interpretations. Our work provides a scalable approach for Chinese ambiguity corpus and insights into LLMs' ambiguity handling, laying a foundation for enhancing Chinese ambiguity research in LLMs. 


\end{abstract}



\maketitle

\section{Introduction}

Linguistic ambiguity is a universal phenomenon across all human languages~\citep{ling_amb}, referring to some sentences that exhibit more than one plausible interpretation. Recognizing and interpreting such ambiguity is important because it helps us avoid misunderstandings and confusion in daily communication. In recent years, Large Language Models (LLMs) have been developing rapidly and becoming popular in intelligent assistants~\citep{Few-shot-learners, deepseekv3, deepseekr1,gemini}. To improve the robustness and trustworthiness of these assistants, the ability to detect ambiguous sentences is essential for preventing losses caused by misunderstandings~\citep{amb_detect_in_requirements, min-etal-2020-ambigqa, bhaskar-etal-2023-benchmarking, wang-etal-2023-know}. 

While most existing research on linguistic ambiguity has focused on English~\citep{ambchatgpt, liu-etal-2023-afraid, mehrabi-etal-2023-resolving, kim-etal-2024-aligning}, comparatively little attention has been paid to Chinese. Due to its unique grammatical features, Chinese is generally considered as more context-dependent than English~\citep{chinese_amb}. Consequently, in the absence of sufficient contextual cues, Chinese sentences or phrases are more prone to multiple plausible interpretations. A relevant work is CHAmbi~\citep{zhang-etal-2024-chambi}, which proposes a natural language inference (NLI) dataset containing Chinese ambiguous sentences. However, most sentences in this dataset are collected from the Internet, rendering it non-scalable. Therefore, exploring methods to systematically generate or expand the corpus of ambiguous sentences is crucial for subsequent research and forms the central focus of this work.

\begin{figure}
    \centering
    \includegraphics[width=0.9\linewidth]{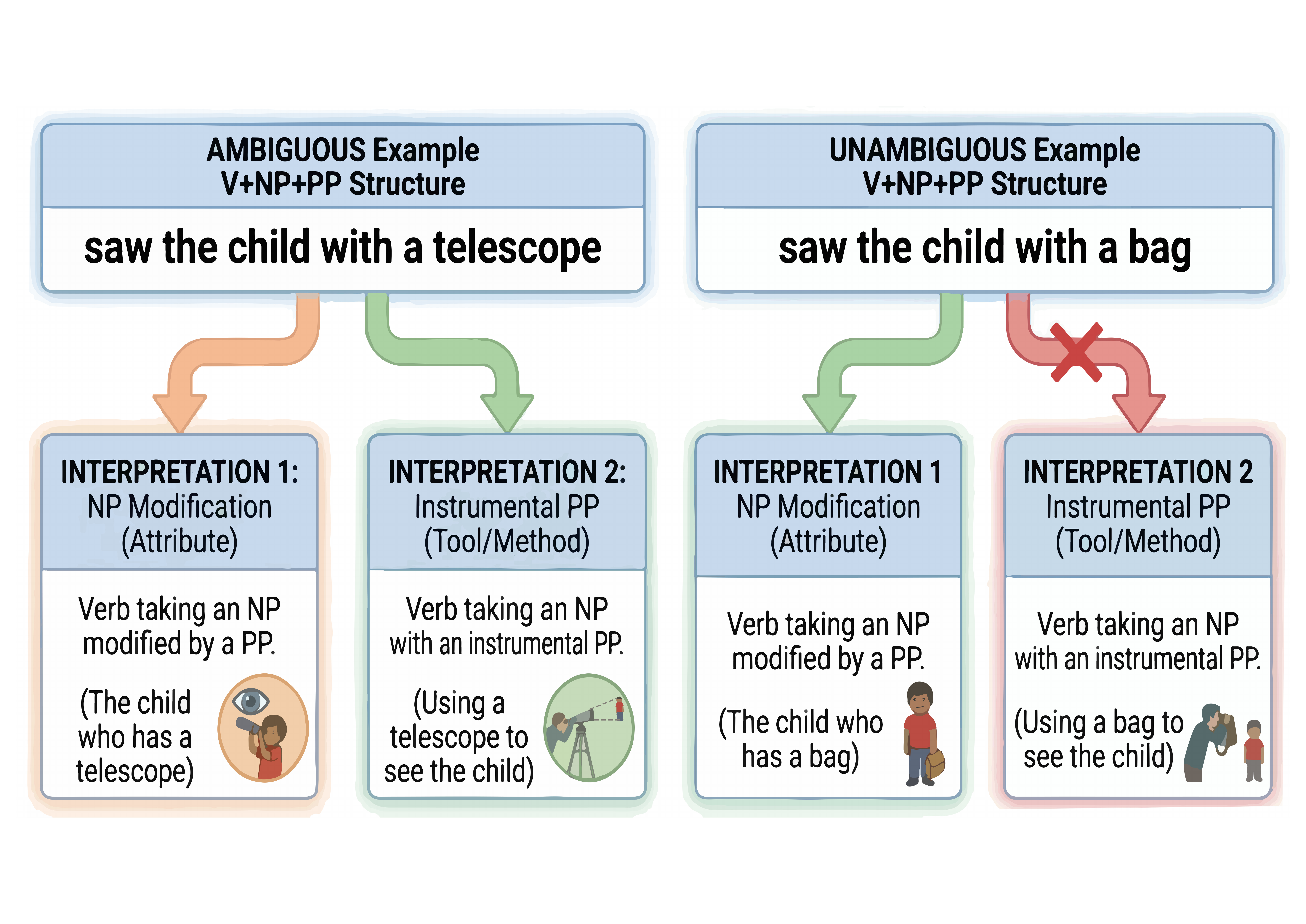}
     \vspace{-35pt}
    \caption{An example of a potential ambiguous structure in PA theory.}
    \label{fig:pa_example}
\end{figure}

Potential Ambiguity Theory (PA Theory)~\citep{pa-theory-part1, pa-theory-part2} posits that certain abstract syntactic configurations, referred to as \emph{potential ambiguous structures}, may give rise to ambiguity upon instantiation when specific conditions are met. For instance (see Figure~\ref{fig:pa_example}), in English, the structure \texttt{V + NP + PP} is likely to lead to ambiguity when the postverbal PP can be interpreted either as modifying the NP or as an adjunct of the verb. Specifically, the phrase “saw the child with a telescope” is ambiguous between two readings: one in which the PP modifies the NP (the child who has a telescope) and another in which it functions as an instrumental modifier of the verb (using a telescope to see the child). By contrast, “saw the child with a bag” is unambiguous, as it fails to trigger the instrumental interpretation. In Chinese, \citet{pa-theory-part2} summarizes 18 types of potential ambiguous structures along with their corresponding ambiguity-triggering conditions. 

Guided by the PA Theory, we design a semi-automatic pipeline. The pipeline collects potential ambiguous structures together with their ambiguity-triggering conditions, then generates and verifies instances via LLMs, followed by human validation. The outcome is the high-quality dataset CHA‑Gen, which contains a total of 5,712 sentences. Among these, 2,414 are ambiguous, while 3,298 are unambiguous. These instances span 18 distinct ambiguity structures. 

Another focus of this work is the systematic evaluation of Chinese ambiguity handling capabilities in LLMs. Based on the proposed CHA-Gen corpus and the existing CHAmbi dataset, we design and carry out assessment using our tailored evaluation methods: direct querying and machine translation. The evaluated models include several general-purpose large language models, including Gemma 3, Qwen 2.5, and Qwen 3, across a range of model scales. The main contributions of this work are summarized as follows:

\begin{enumerate}
    \item We develop a semi-automatic pipeline to construct a large corpus of Chinese sentences with structural ambiguity, CHA-Gen\footnote{\url{https://github.com/SpaJune/CHA-Gen}}. To the best of our knowledge, CHA-Gen is the first Chinese ambiguity dataset grounded in PA Theory. It enriches available resources for research on Chinese linguistic ambiguity. 

    \item We conduct ambiguity identification and comparison tasks to evaluate LLMs' ability of ambiguity detection and discrimination. LLMs still face challenges in such tasks, as shown in previous work~\citep{liu-etal-2023-afraid, zhang-etal-2024-chambi}, while chain-of-thought (CoT) prompting can significantly improve the performance in most cases. We analyze their generated rationale and provide insights for future study. 

    \item We investigate LLMs’ uncertainty in Chinese-to-English translation using semantic entropy as an uncertainty quantification metric. The results show that LLMs have higher generation uncertainty for ambiguous than unambiguous sentences. Moreover, Base models capture semantic diversity in certain ambiguous cases, whereas instruction tuning reduces uncertainty and causes overconfident predictions. A further case study investigates how models are biased toward dominant interpretations. These findings highlight potential limitations of instruction tuning and raise concerns about misunderstandings in cross-lingual communication. 

\end{enumerate}

The remainder of the article is organized as follows. We review related work in Section \ref{sec2} and introduce CHA-Gen Corpus in Section \ref{sec3}. The details of the evaluation via direct querying and machine translation are presented in Section \ref{sec4} and Section \ref{sec5}, respectively. Section \ref{sec6} summarizes the article.

\section{Related Work}
\label{sec2}

Research efforts have been devoted to constructing relevant datasets and benchmarks, as well as investigating how LLMs behave when confronted with ambiguous inputs. We review related work on these topics.

\textbf{Datasets and benchmarks of ambiguity.} Previous research has released datasets and benchmarks focusing on visual ambiguity \citep{VQA,chen-etal-2024-unified-hallucination} and vision-based context ambiguity~\citep{ma-etal-2024-3am,MUCAR}. \citet{wildenburg-etal-2024-pre} introduce the DUST dataset, which comprises underspecified sentences, and conduct experiments to assess whether LLMs can detect syntactic ambiguity based on perplexity. However, although significant progress has been made in disambiguation for English and the multimodal domain, available resources for Chinese remain relatively limited. In the Chinese context, \citet{he-etal-2020-box} present a dataset that encompasses both lexical and syntactic ambiguities, using it to evaluate the commonsense reasoning abilities of LLMs. More recently, \citet{zhang-etal-2024-chambi} constructs CHAmbi, a fine-grained Chinese benchmark specifically designed for ambiguity detection and resolution. However, the sentences in CHAmbi are primarily collected from the Internet, which limits the scalability and practical application. Different from CHAmbi, we develop a novel process to construct ambiguous Chinese sentences/phrases based on potential ambiguous structures. This way enhances the scalability of corpus construction while maintaining linguistic quality.

\textbf{Inspecting LLMs' Behavior in Handling Ambiguity.} LLMs' ambiguity-handling abilities have been evaluated primarily through direct querying and behavior-based analysis. 

The idea of direct querying is straightforward. LLM models are explicitly prompted to identify whether a given sentence is ambiguous \citep{zhang-etal-2024-chambi, ambchatgpt}. In particular, \citet{amb_kdd25} conducts a comprehensive study of different prompting strategies. Another line of direct querying research adopts a comparative method \citep{wildenburg-etal-2024-pre}, in which models are given pairs of ambiguous and unambiguous sentences to identify the more underspecified one.

To comprehensively understand LLMs' ability to understand ambiguity, examining how they handle ambiguous sentences across different downstream tasks is also important. \citet{liu-etal-2023-afraid,zhang-etal-2024-chambi} adopt multilabel NLI tasks in order to see whether LLMs can be aware of the uncertainty of ambiguous inputs. In addition, \citet{liu-etal-2023-afraid} evaluates ambiguity awareness in language models by comparing continuation distributions conditioned on ambiguous sentences and their disambiguations. Using sampled textual continuations, they assess whether continuations from valid interpretations are less surprising under the ambiguous context than those from distractor contexts. However, these experiments require a predefined enumeration of multiple plausible disambiguations. 

\citet{mehrparvar-pezzelle-2024-detecting} employs machine translation to detect sentence ambiguity by measuring discrepancies between back-translations and the original sentences in representation space. The underlying intuition is that an ambiguous sentence has multiple interpretations and thus leads to varying plausible translations with different semantics. Inspired by this insight, we explore LLMs’ translation uncertainty behavior between ambiguous and unambiguous sentences. This allows us to study ambiguity-related uncertainty without explicitly enumerating alternative hypotheses, while also revealing which readings LLMs favor when ambiguity is left underspecified.

In summary, to enable a more comprehensive evaluation, we adopt both the direct querying (including the comparative method) and the behavior-based analysis with machine translation. It is worth noting that machine translation is employed to probe LLMs’ uncertainty behavior in this work rather than to detect ambiguous sentences. We also believe that, compared to direct querying, translation tasks facilitate a deeper understanding of LLMs' implicit interpretations of ambiguous inputs. 

\section{CHA-Gen Corpus}
\label{sec3}



To systematically investigate the ability of LLMs in Chinese ambiguity identification and understanding, it is essential to construct a corpus that is large-scale, well-categorized, and finely annotated.
However, existing resources remain insufficient for this purpose. To fill this gap, this paper presents the Chinese Ambiguity-Generation (CHA-Gen) corpus. The following subsections elaborate on its construction process and corpus composition.

\begin{figure}
    \centering
    \includegraphics[width=0.9\linewidth]{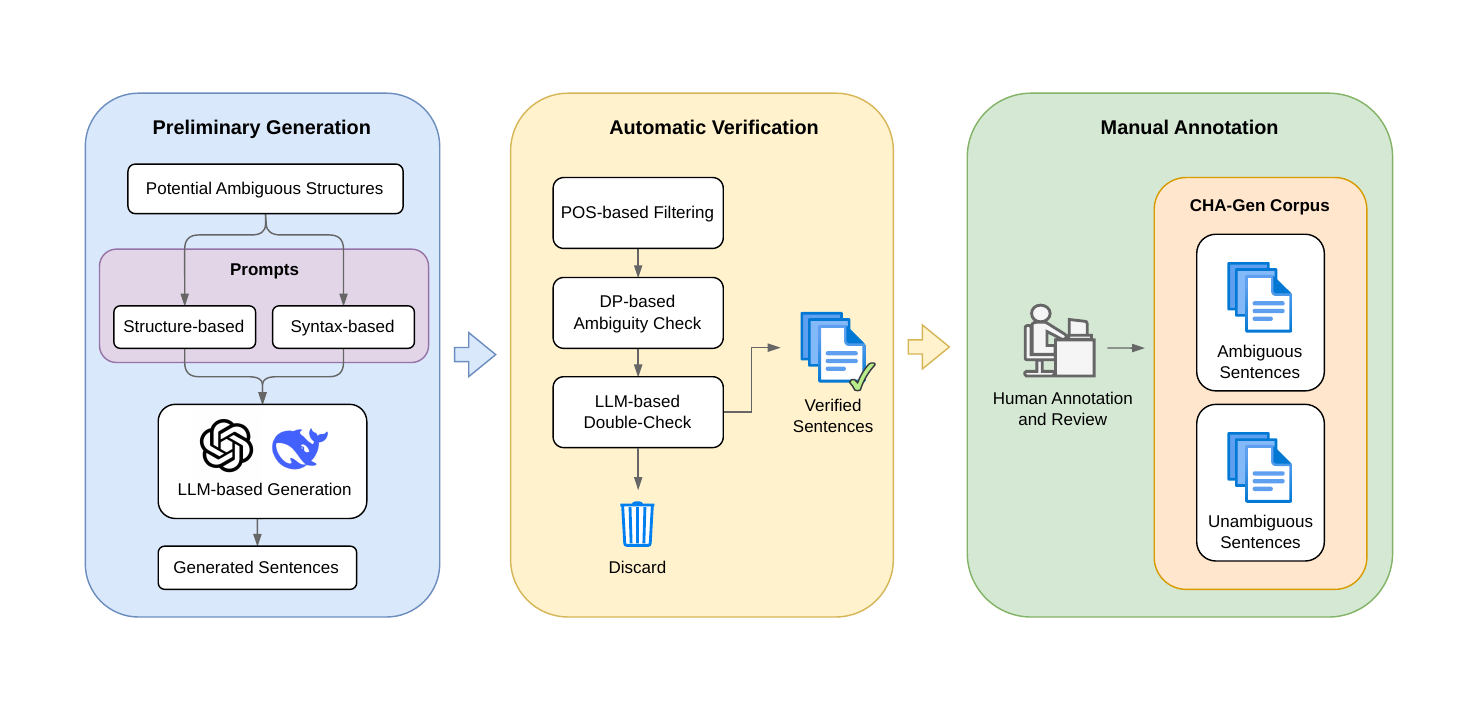}
    \caption{The Pipeline for CHA-Gen Corpus Construction}
    \label{fig:corpus_construction_pipeline}
\end{figure}

\subsection{Corpus Construction}
\label{Corpus Construction}

We devise a semi-automatic pipeline to construct a dedicated Chinese ambiguity corpus. The procedure involves three key stages: preliminary generation, automatic verification and manual annotation, as shown in Figure~\ref{fig:corpus_construction_pipeline}.

\begin{table}[htbp]
    \centering
    \caption{The number of ambiguous/unambiguous sentences in CHA-Gen. Abbreviations: VP (Verb Phrase), NP (Noun Phrase), V (Verb), N (Noun), ADJ (Adjective), Q (Quantifier) and PREP (Preposition).}  
    \label{tab:ca-gen}
    \small

        \begin{tabular}{c l c c}  
        \toprule
        No. & Potential Ambiguous Structure & Ambiguous & Unambiguous \\  
        \midrule
        1  & VP+的是+NP               & 356 & 234 \\  
        2  & Q+NP1+的+NP2             & 243 & 362 \\
        3  & N1+N2+N3                 & 220 & 545 \\
        4  & ADJ+N1+N2                & 72  & 50  \\
        5  & V1+V2+NP                 & 155 & 132 \\
        6  & N+V                      & 49  & 40  \\
        7  & N1+的+N2+和+N3           & 47  & 103 \\
        8  & N+V+NP+AP                & 163 & 254 \\
        9  & 全部+VP+的+NP            & 307 & 130 \\
        10 & N1+N2                    & 25  & 22  \\
        11 & VP+ADJ+的+N              & 61  & 67  \\
        12 & N1+和+N2+的+N3           & 69  & 102 \\
        13 & VP+Q+NP                  & 33  & 653 \\
        14 & V+ADJ+N                  & 50  & 137 \\
        15 & VP1+VP2+的+N             & 158 & 163 \\
        16 & V+N                      & 29  & 21  \\
        17 & VP+N1+的+N2              & 114 & 113 \\
        18 & PREP+N1+的+N2            & 263 & 170 \\
        \midrule
            & Total                    & 2,414 & 3,298 \\
        \bottomrule
    \end{tabular}
 
\end{table}

\textbf{Preliminary Generation.} This stage aims to establish a typology of ambiguous structures and generate an initial, scalable corpus. Following the PA Theory, we first identify 18 types of potential ambiguous structures. For each type, a small set of ambiguous sentences is manually constructed by combining brainstorming with the collection of examples from online resources. Building upon these human-curated instances, we then leverage the pattern-generation capabilities of LLMs \citep{Few-shot-learners} to produce a large number of sentences with analogous structures and potential ambiguity. Specifically, to enhance generation efficiency and diversity, we design two types of prompts:

\begin{itemize}
\item \textbf{Structure-based Prompts.}
These prompts instruct the LLMs to generate syntactically ambiguous sentences strictly according to potential ambiguous structures.

\item \textbf{Syntax-based Prompts.} 
These prompts leverage syntactic flexibility and contextual constraints to guide LLMs in constructing ambiguous sentences.

\end{itemize}

Note that the former prompts adhere to defined structural rules, whereas the latter prompts rely on syntactic relationships. More details can be found in Appendix \ref{prompt_template}.

\textbf{Automatic Verification.} Since automatically generated sentences may deviate from the target ambiguous structures, this stage aims to automatically verify their structural consistency. We leverage HanLP~\footnote{https://www.hanlp.com/} for Chinese word segmentation, Part-of-Speech (POS) tagging and Dependency Parsing (DP), supplemented by DeepSeek-R1~\citep{deepseekr1} for double-checking. The verification consists of three folds:

\begin{itemize}
\item \textbf{POS-based Filtering.}
Sentences showing significant discrepancies in word count or POS tag distribution against the predefined ambiguous structure are filtered.

\item \textbf{DP-based Ambiguity Check.} For the remaining sentences, we insert specific words and analyze whether the root node of their dependency tree changes. Sentences with altered root nodes are classified as ambiguous, others as unambiguous.

\item \textbf{LLM‑based Double‑Check.} Filtered sentences, along with their corresponding syntactic structures and explanations of potential ambiguity, are provided to DeepSeek-R1. The model is prompted to determine whether ambiguity exists and to provide corresponding reasoning (see Appendix \ref{validation_prompt} for details). To improve efficiency in eliciting well-structured evaluations from the LLM, we adopt the LLM-Eval evaluation method~\citep{lin-chen-2023-llm}.

\end{itemize}

\textbf{Manual Annotation.}
 In this stage, the instances labeled as Ambiguous by DeepSeek-R1 are further manually reviewed. Two annotators independently annotate the filtered sentences and conduct cross-validation. To improve annotation quality, the annotators inspect and refine the rationales for ambiguity or non-ambiguity (initially provided by the model or in draft form), ensuring descriptions are accurate and consistent.

\subsection{Corpus Composition and Statistics}

Through the aforementioned construction pipeline, we obtain the Chinese Ambiguity-Generation (CHA-Gen) corpus. This corpus contains 2,414 ambiguous and 3,298 unambiguous sentences, covering 18 structural ambiguity types. For each ambiguous structure, the corpus includes both ambiguous instances that satisfy the ambiguity-triggering conditions and unambiguous counterparts that share similar surface forms but admit a single interpretation. The overall statistics are summarized in Table~\ref{tab:ca-gen}. Each ambiguous sentence in CHA-Gen is associated with its specific ambiguous
structure and the corresponding linguistic explanation. More details please refer to Appendix \ref{CHA-Gen_ambiguity_structure}.

\section{Evaluation of LLMs via Direct Querying}
\label{sec4}


We adopt the direct querying method to evaluate the ability of LLMs for ambiguity detection and discrimination. Following the experimental settings of related work~\citep{zhang-etal-2024-chambi, wildenburg-etal-2024-pre}, we design two query tasks: ambiguity identification and ambiguity comparison. To enhance the reliability and generality of evaluation, we also incorporate the CHAmbi dataset for validation.


\begin{table*}[htbp]
\centering
\caption{Prompts for LLMs. Chinese prompts are used in our experiments, and English versions are provided for reference.}
\label{tab:direct-prompt}
\small

\resizebox{0.9\textwidth}{!}{%
\begin{tabular}{@{}l p{0.75\textwidth}@{}}
\toprule
Prompt Strategies & Prompt \\
\midrule
Direct Prompt for Identification Task &
请判断下面的句子/词组是否具有歧义。只回答“是”或“否”。

句子：\texttt{\{sentence\}}

\textit{Please determine whether the following sentence/phrase is ambiguous.
Answer only "Yes" or "No".}

\textit{Sentence:}  \texttt{\{sentence\}}\\[0.6em]
\midrule
Direct Prompt for Comparison Task&
下面有两个句子/词组，请判断哪个更容易引起歧义。只回答“1”或“2”。

句子1：\texttt{\{sent\_1\}}

句子2：\texttt{\{sent\_2\}}

\textit{Below are two sentences/phrases. Please determine which one is more likely
to cause ambiguity. Answer only "1" or "2".}

\textit{Sentence 1: }\texttt{\{sent\_1\}}

\textit{Sentence 2: }\texttt{\{sent\_2\}}
\\

\midrule
CoT Prompt for Identification Task&
请判断下面的句子/词组是否具有歧义。请逐步思考，写出你的思考过程，在最后一行单独写出结论“是”或“否”。

句子：\texttt{\{sentence\}}

\textit{Please determine whether the following sentence/phrase is ambiguous.
Please reason step by step and write out your reasoning process.
In the final line, provide the conclusion only: "Yes" or "No".}

\textit{Sentence: } \texttt{\{sentence\}}

\\

\midrule
CoT Prompt for Comparison Task&

下面有两个句子/词组，请判断哪个更容易引起歧义。请逐步思考，写出你的思考过程，在最后一行单独写出结论“1”或“2”。

句子1：\texttt{\{sent\_1\}}

句子2：\texttt{\{sent\_2\}}

\textit{Below are two sentences/phrases. Please determine which one is more likely
to cause ambiguity. Please reason step by step and write out your reasoning process.
In the final line, provide the conclusion only: "1" or "2".}

\textit{Sentence 1: }\texttt{\{sent\_1\}}

\textit{Sentence 2: }\texttt{\{sent\_2\}}

\\

\bottomrule
\end{tabular}
}
\end{table*}

\subsection{Evaluation Setup}
\label{Evaluation Setup}
The formal settings and evaluation procedures for the query tasks are specified below. 

\textbf{Query Task 1: Ambiguity Identification.} 
Given a single input sentence, LLMs are required to determine whether the sentence is ambiguous. We employ two prompting strategies (see Table~\ref{tab:direct-prompt}): a Direct Prompt that instructs the model to output only “Yes” or “No”, and a Chain-of-Thought (CoT) Prompt that elicits step-by-step reasoning before a final “Yes” or “No” decision. We record the generation probabilities of 是 (Yes) and 否 (No) at the first token position.
The prediction is the option with the highest probability, and we evaluate the performance using Accuracy and Macro-F1 metrics. The statistics of the benchmark datasets are summarized in Table~\ref{tab:stat_data_identification}.


\textbf{Query Task 2: Ambiguity Comparison.} 
Given a pair of sentences, LLMs are required to identify which one is relatively more ambiguous. We employ two prompting strategies (see Table~\ref{tab:direct-prompt}): a Direct Prompt that instructs the model to output “1” or “2” to indicate the more ambiguous sentence, and a CoT Prompt that guides it to conduct step-by-step reasoning before concluding with “1” or “2”. We construct the sentence pairs as follows: For each ambiguous sentence in the CHA-Gen dataset, we select a corresponding unambiguous counterpart based on surface similarity. For the CHAmbi dataset, we directly use its provided unambiguous counterpart to construct sentence pairs. In addition, to mitigate order effects~\cite{multichoice_robust}, each sentence pair is queried twice with their positions swapped. This ensures that options “1” and “2” appear equally in the comparison task, effectively balancing the number of ambiguous and unambiguous sentences. Evaluation performance is also evaluated using Accuracy and Macro‑F1. The statistics of the benchmark datasets are presented in Table \ref{tab:stat_data_comparison}.




\begin{table}[]

\caption{Statistics of benchmark datasets}

\begin{subtable}{0.48\textwidth}
\small
\centering
\caption{Ambiguity identification task}
\label{tab:stat_data_identification}
\begin{tabular}{@{}c cc@{}}
\toprule
               & CHA-Gen & CHAmbi \\ \midrule
\# Amb. Sent   &         2,414&        893\\
\# Unamb. Sent &         3,298&        1,784\\
Total &      5,712   &  2,677      \\
Avg. Len.      &         7.38&        19.06\\ \bottomrule
\end{tabular}
\end{subtable}
\hspace{0.01\textwidth}
\begin{subtable}{0.48\textwidth}
\small
\centering
\caption{Ambiguity comparison task}
\label{tab:stat_data_comparison}
\begin{tabular}{@{}c cc@{}}
\toprule
                      & CHA-Gen & CHAmbi \\ \midrule
\# Pairs              &         2,414&        1,784\\
\# Unique Amb. Sent   &         2,414&        877\\
\# Unique Unamb. Sent &         856&        1,772\\
Avg. Len.             &         7.55&        18.42\\ \bottomrule
\end{tabular}

\end{subtable}
\end{table}

\begin{table*}[t]
\small
\centering
\caption{Results on ambiguity identification and ambiguity comparison tasks across two datasets.}
\label{tab:identification_results}
\resizebox{\textwidth}{!}{
\begin{tabular}{l cc cc cc cc}
\toprule
\multirow{3}{*}{Model}
& \multicolumn{4}{c}{Ambiguity Identification Task}
& \multicolumn{4}{c}{Ambiguity Comparison Task} \\
\cmidrule(lr){2-5} \cmidrule(lr){6-9}
& \multicolumn{2}{c}{CHA-Gen}
& \multicolumn{2}{c}{CHAmbi}
& \multicolumn{2}{c}{CHA-Gen}
& \multicolumn{2}{c}{CHAmbi} \\
\cmidrule(lr){2-3} \cmidrule(lr){4-5}
\cmidrule(lr){6-7} \cmidrule(lr){8-9}
& Acc $\uparrow$ & Macro-F1 $\uparrow$
& Acc $\uparrow$ & Macro-F1 $\uparrow$
& Acc $\uparrow$ & Macro-F1 $\uparrow$
& Acc $\uparrow$ & Macro-F1 $\uparrow$ \\
\midrule
Gemma3-1B-IT & 0.5285 & 0.5260 & 0.3463 & 0.2976 & 0.5058 & 0.4227 & 0.2618 & 0.2603 \\
Gemma3-4B-IT & 0.4251 & 0.3028 & 0.3526 & 0.3247 & 0.5501 & 0.5501 & 0.4271 & 0.3372 \\
Gemma3-12B-IT & 0.4540 & 0.3659 & 0.3407 & 0.3210 & 0.5164 & 0.4156 & 0.4268 & 0.3834 \\
Gemma3-27B-IT & 0.5011 & 0.4560 & 0.3635 & 0.3632 & 0.5379 & 0.5011 & 0.4285 & 0.4234 \\
\midrule
Qwen2.5-3B-Instruct & 0.5826 & 0.3948 & 0.6362 & 0.3965 & 0.5414 & 0.5385 & 0.5605 & 0.5603 \\
Qwen2.5-7B-Instruct & 0.5532 & 0.5055 & 0.6081 & 0.4454 & 0.5151 & 0.3876 & 0.3814 & 0.3779 \\
Qwen2.5-14B-Instruct & 0.6266 & 0.5906 & 0.5570 & 0.4802 & 0.5217 & 0.4235 & 0.6244 & 0.6105 \\
Qwen2.5-32B-Instruct & 0.5004 & 0.4536 & 0.4531 & 0.4485 & 0.5379 & 0.4945 & 0.5048 & 0.4946 \\
\midrule
Qwen3-4B & 0.4265 & 0.3048 & 0.3586 & 0.3576 & 0.5543 & 0.5542 & 0.4367 & 0.3739 \\
Qwen3-8B & 0.4772 & 0.4306 & 0.4225 & 0.4140 & 0.5485 & 0.5355 & 0.4120 & 0.3585 \\
Qwen3-14B & 0.4793 & 0.4149 & 0.4285 & 0.4221 & 0.5580 & 0.4963 & 0.4490 & 0.4361 \\
Qwen3-32B & 0.5158 & 0.4819 & 0.4991 & 0.4746 & 0.5466 & 0.4857 & 0.3876 & 0.3876 \\
\bottomrule
\end{tabular}
}
\end{table*}

\begin{figure}
    \centering

    \begin{subfigure}{\linewidth}
        \centering
        \includegraphics[width=0.75\linewidth]{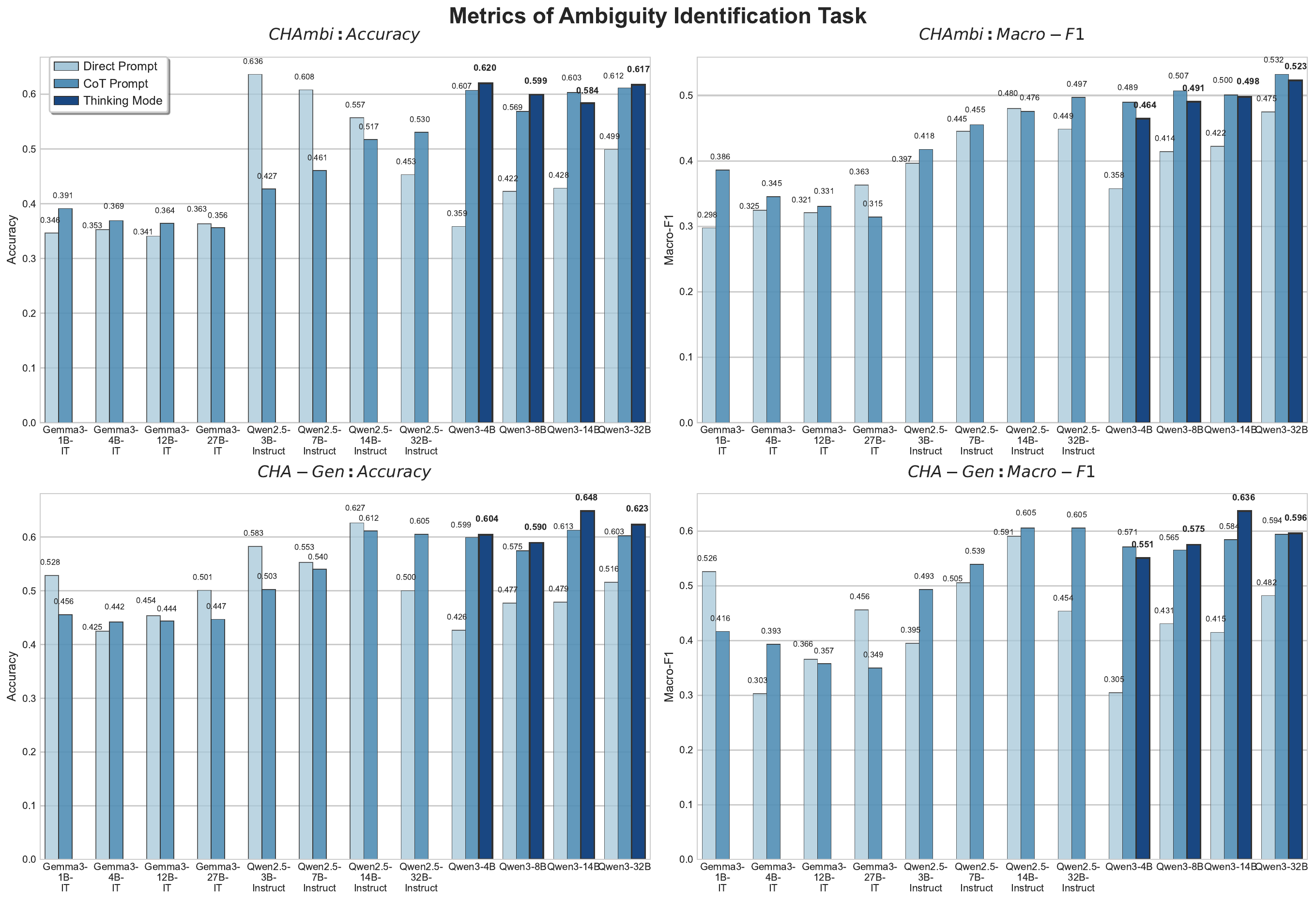}
        \caption{Accuracy and Macro-F1 in ambiguity identification task.}
        \label{fig:identification_cot}
    \end{subfigure}
    \vfill
    \begin{subfigure}{\linewidth}
        \centering
        \includegraphics[width=0.75\linewidth]{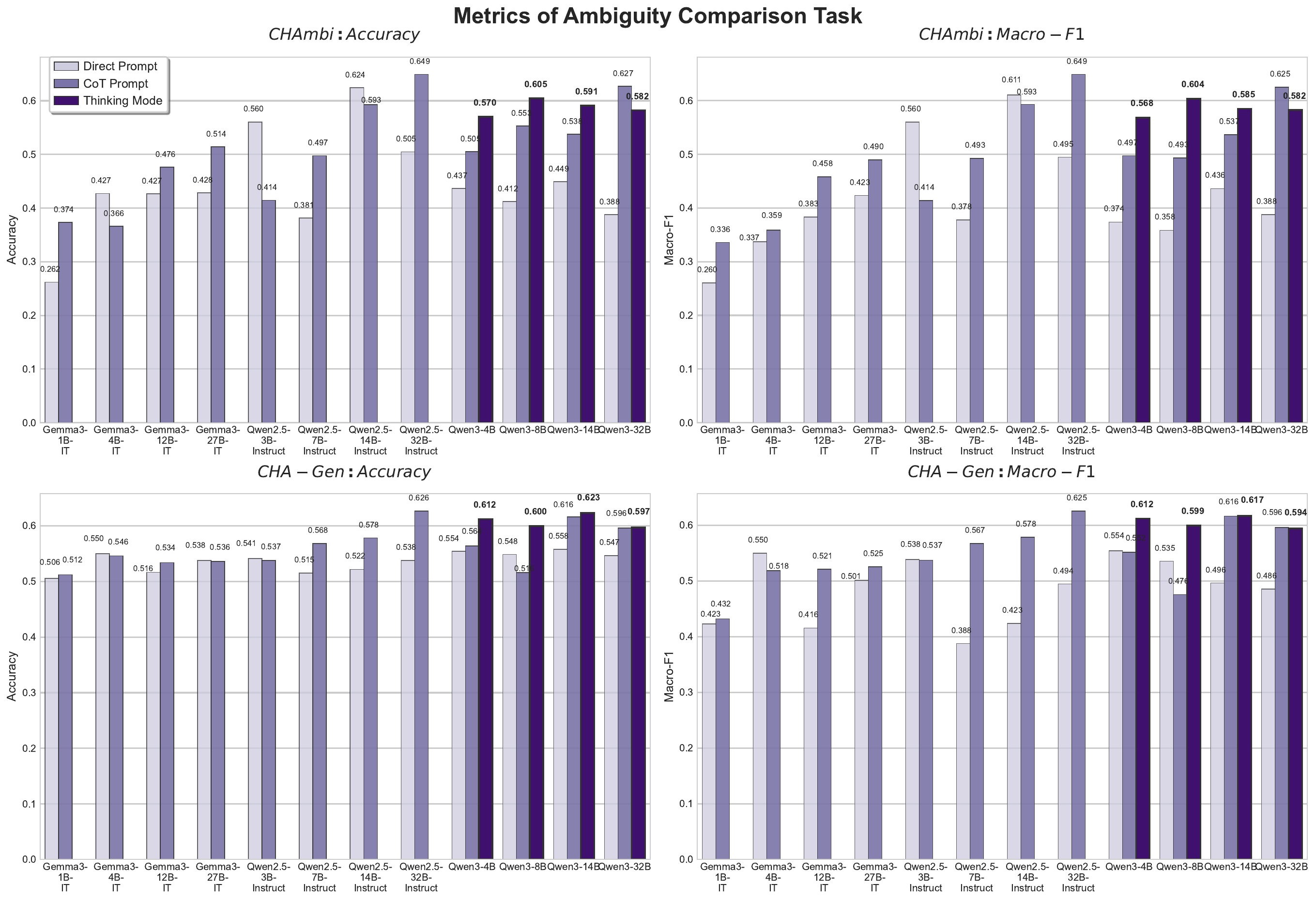}
        \caption{Accuracy and Macro-F1 in ambiguity comparison task.}
        \label{fig:comparison_cot}
    \end{subfigure}

    \caption{Overall performance comparison under various prompts and thinking modes.}
    \label{fig:cot_metrics}
\end{figure}

\subsection{Evaluation Results and Discussion}
The evaluation results are presented in Table~\ref{tab:identification_results}. In the ambiguity identification task, Qwen2.5-14B-Instruct achieves the highest Macro-F1 score among all evaluated models across both datasets. In the ambiguity comparison task on CHA-Gen, Gemma3-4B-IT and Qwen3-4B achieve the best Macro-F1, followed by Qwen2.5-3B-Instruct and Qwen3-8B. On CHAmbi, Qwen2.5-14B-Instruct again demonstrates higher performance, confirming its robustness across both tasks. 

Notably, the overall performance across all models and datasets is relatively low (around 0.5), with Accuracy and Macro-F1 scores often dropping even further for most dataset-model combinations. This indicates that many models perform at a level comparable to random guessing on the proposed tasks. Importantly, factors (larger model sizes or higher version iterations) typically associated with enhanced general language capabilities, such as Qwen2.5 $\rightarrow$ Qwen3, do not necessarily translate to better performance, highlighting the unique challenges of ambiguity detection and discrimination.


\textbf{Effects of Explicit Reasoning and CoT Prompt.} This part aims to evaluate how explicit reasoning and CoT Prompt shape model behavior and performance. To implement the explicit reasoning setting, we take advantage of the configurable thinking mode in Qwen3, which can be flexibly enabled or disabled. The performance changes induced on both tasks are reported separately in Figure~\ref{fig:identification_cot} and Figure~\ref{fig:comparison_cot}. For the ambiguity identification task, most models achieve higher Macro-F1 under CoT prompt, albeit with a concurrent decrease in overall Accuracy. This observation suggests that CoT prompt encourages models to produce more balanced predictions. For the ambiguity comparison task, where ground-truth labels (1 and 2) are balanced, Accuracy and Macro-F1 exhibit consistent trends under CoT prompting, further validating the effectiveness of explicit reasoning for ambiguity judgments.

\begin{figure}
    \centering
    \includegraphics[width=0.9\linewidth]{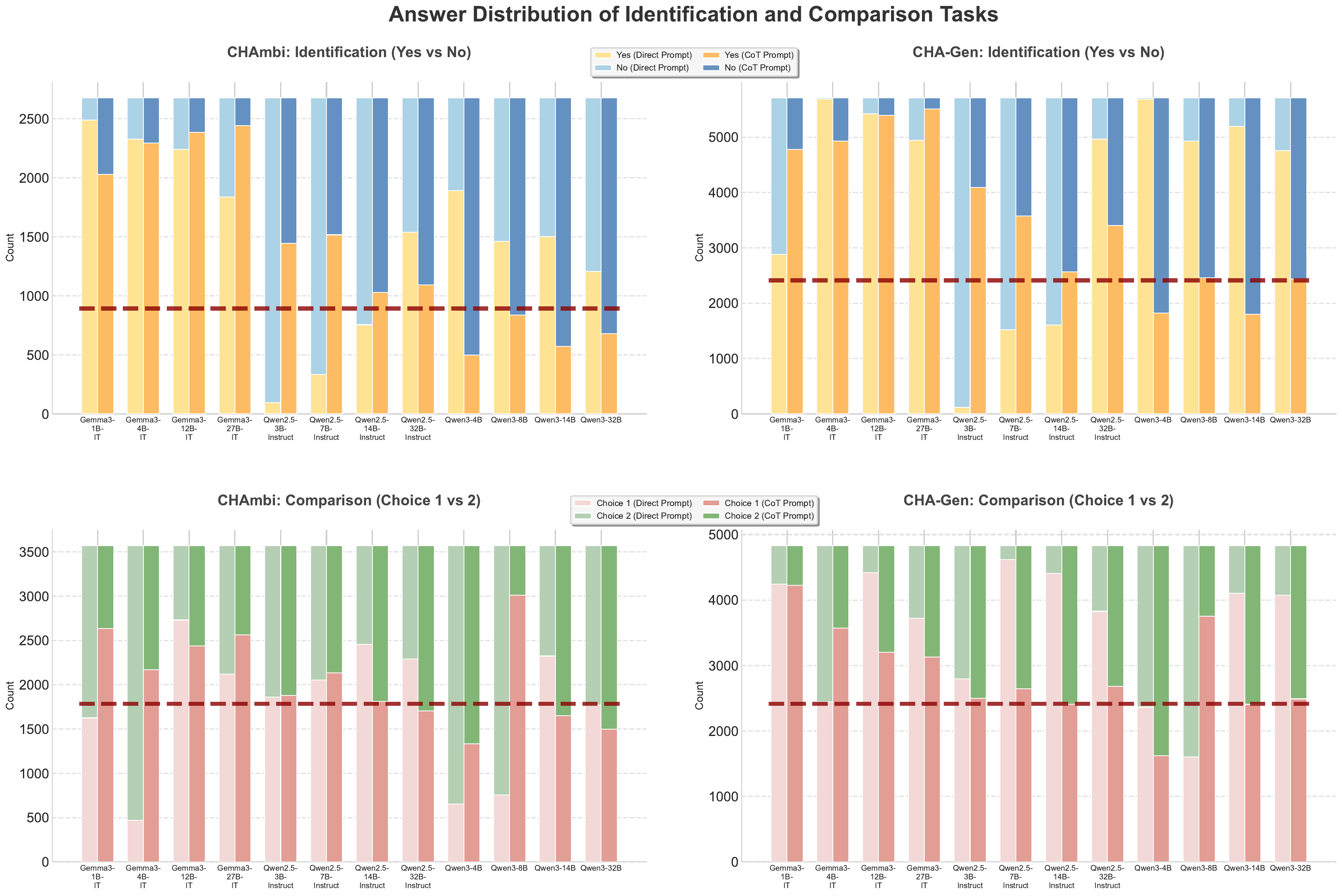}
    \caption{The answer distributions of CHAmbi and CHA-Gen in ambiguity identification and comparison tasks, with the dashed line indicating the optimal balance point.}
    \label{fig:ans_distribution}
\end{figure}

\textbf{Answer Distributions.} To delve into the observed performance trends, we analyze the answer distribution of each model, which is visualized in Figure~\ref{fig:ans_distribution}. We observe that many models exhibit strong biases towards specific answers (e.g., consistently favoring “Yes” or “1”) in the absence of explicit reasoning and CoT prompts. This phenomenon is particularly pronounced in Gemma3-based models. Such systematic biases, independent of the actual input content, indicate that many models struggle to perform reliable single-token judgments. In contrast, when reasoning is enabled or a CoT prompt is applied, answer distributions become more balanced in most cases, suggesting that LLMs effectively leverage relevant domain knowledge through explicit reasoning. 

Overall, forming such judgments via single-word responses poses inherent difficulties for modern LLMs. Directly instructing LLMs to provide a one-word decision for ambiguity assessment, as done in prior work~\citep{zhang-etal-2024-chambi}, may not constitute a reliable evaluation protocol and can underestimate their true capabilities. Nevertheless, our results show that ambiguity detection and discrimination remain highly challenging even when explicit reasoning is enabled. This observation motivates us to further analyze the models’ generated rationales to gain deeper insights into their ambiguity detection processes.

\subsection{Further Analysis of Rationales in CoT Prompting}

We further examine the rationales generated by Qwen3-32B to gain a deeper understanding. Several examples from CHA-Gen are provided in Table~\ref{tab:cot_rationale_errors}. The typical failure cases are summarized as follows:
\begin{itemize}
    \item \textbf{Ambiguity Blindness.} The most typical failure stems from an insufficient sensitivity to potential sources of ambiguity, where the model fails to recognize latent ambiguities in the sentence. In this case, the model prematurely concludes that the sentence is unambiguous, without sufficiently exploring alternative interpretations (See sample 5 in Table~\ref{tab:cot_rationale_errors}). 
    \item  \textbf{Ambiguity Misattribution.} Another failure arises from a mismatch between the model’s interpretation of ambiguity and that of humans. In such cases, the model correctly identifies certain linguistic phenomena (e.g., subject omission or nominalization) but incorrectly treats them as sufficient evidence of ambiguity. As a result, it reasons in an inappropriate direction, constructing multiple implausible interpretations and leading to misguided judgments, which frequently happens even when the final prediction turns out to be correct (See Samples~1–3 in Table~\ref{tab:cot_rationale_errors}). This highlights a fundamental discrepancy between human and LLMs in their conceptualization of ambiguity. 
    \item \textbf{Premature Ambiguity Resolution.} In some cases, Qwen3-32B correctly identifies the presence of ambiguity and generates multiple plausible interpretations. However, before making the final decision, it erroneously dismisses one interpretation as less natural or less likely in real-world usage, thereby prematurely collapsing the ambiguity and producing an incorrect, unambiguous judgment (See Sample 4 in Table~\ref{tab:cot_rationale_errors}). 
\end{itemize}



Based on these observations, improving LLM performance in recognizing ambiguous sentences requires enhanced linguistic understanding and more robust reasoning capabilities. Deeper, more systematic reasoning may help models better perceive and maintain multiple interpretations. Additionally, addressing Ambiguity Misattribution and Premature Ambiguity Resolution may benefit from a more precise and nuanced definition of “ambiguity,” aligning model reasoning more closely with human judgments.

\begin{table}[t]
\caption{Examples of incorrect CoT rationales in the ambiguity detection task. The shown rationales are translated into English.}
\label{tab:cot_rationale_errors}
\centering
\small
\resizebox{0.9\textwidth}{!}{%
\begin{tabular}{l>{\raggedright\arraybackslash}
  p{4.6cm}
  >{\raggedright\arraybackslash}p{2.6cm}
  >{\raggedright\arraybackslash}p{6.0cm}
  l
}
\hline
 Sample ID&Sentences & Model Prediction &Generated Rationale (Excerpt)  &Error\\
\hline

 1&Sent1: 信任的是少数几位朋友。\par
\emph{(The entity being trusted / trust (sth.) is a few friends.)}\par
Sent2: 相信的是科学证据。\par
\emph{(Belief is placed in scientific evidence.)}
&
Qwen3-32B: Sent2 is more ambiguous. (Incorrect)
&“Scientific evidence” is a relatively abstract and broad concept, and different individuals may interpret it differently [...] prone to varying interpretations among different people, leaving considerable room for misinterpretation.
 &Ambiguity Misattribution: Lexical\\
\midrule

 2&Sent1: 关注的是他们。 \par
\emph{(The ones being focused on/ focus on (sth.) are them. )} \par
Sent2: 被期待的是他们。\par
\emph{(The ones being expected are them. )} \par
&
Qwen3-32B: Sent2 is more ambiguous (Incorrect)
&
The verb “expect” carries emotional connotations, while the nominalized structure “the expected (one)” easily prompts questions like “Who expects?” and “What is expected?”, thereby increasing the likelihood of ambiguity.
&
Ambiguity Misattribution: Pragmatic\\

\midrule

 3&Sent1: 喜欢的是小明。 \par
\emph{(The one being liked/ like (sth.) is Xiaoming. )} \par
Sent2: 喜欢的电影类型是喜剧。\par
\emph{((My) Favorite movie genre is comedy.)} \par
&
Qwen3-32B: Sent1 is more ambiguous. (Correct)
&
There are several possible interpretations, such as: - I like Xiaoming. - He likes Xiaoming. - Everyone likes Xiaoming. [...]
**Sentence 1** Because the subject is missing, it can easily lead to misunderstanding about “who likes Xiaoming,” thus **making it more prone to ambiguity**.
&
Ambiguity Misattribution: Referential\\

\midrule

 4 &Sentence: 全部损坏的机器设备 \par
 \emph{(All damaged machinery and equipment)} \par
 \emph{(Completely damaged machinery and equipment(s))}
 & 
  Qwen3-32B: The sentence is unambiguous. (Incorrect)
 & 
 [...] \par
- Interpretation 1: “All” modifies “damaged machinery and equipment,” meaning “all machinery and equipment are damaged.” \par
- Interpretation 2: “All” modifies “damaged,” indicating “damaged to a very thorough extent,” i.e., “the machinery and equipment are completely destroyed.” \par
[...] \par
While two interpretations exist, Chinese grammatical conventions suggest “all” more likely modifies the entire “damaged machinery and equipment,” meaning “all damaged machinery and equipment.” \par
[...] \par
Conclusion: No

 &
 Premature Ambiguity Resolution
 \\

\midrule

 5 &Sentence: 照顾花园的园丁 \par
 \emph{(The gardener who tends the garden)} \par
 \emph{(Take care of the gardener of the garden)}
 & 
  Qwen3-32B: The sentence is unambiguous. (Incorrect)
 & 
 [...] 
 1. Sentence Structure Analysis: \par
 - The subject is “gardener.” \par
 - “Tends the garden” is a modifying phrase describing the gardener's duties or characteristics. \par
 [...] 
 Here, “tending the garden” modifies “gardener” with no alternative possible interpretation.  [...] No alternative interpretations exist.

 &
Ambiguity Blindness
 \\

 \bottomrule
\end{tabular}
}

\end{table}

\section{Evaluation of LLMs via Machine Translation}
\label{sec5}

Simply querying LLMs to recognize ambiguity is insufficient to assess whether they really perceive an input as ambiguous. Instead, it is necessary to examine models' behavior when confronted with ambiguous inputs, specifically whether they can exhibit uncertainty. We focus on the Chinese-to-English translation task. Ambiguous source sentences may admit multiple plausible interpretations. Accordingly, we hypothesize that ambiguous sentences give rise to multiple valid translations with greater semantic diversity. Our goal is to measure each model’s sensitivity to ambiguity by quantifying the uncertainty reflected in its subsequent outputs.

\subsection{Evaluation Setup}

For each sentence, we prompt the models to sample 50 translations, forming a set of translations $T$. The instructions used for sampling translations are provided in Appendix~\ref{apd:sample_trans}. We assess the semantic uncertainty of $T$ to determine whether it can cover diverse meanings using uncertainty quantification (UQ). Concretely, the semantic entropy proposed by \citep{semantic_set} is adopted to measure uncertainty. Formally, given $T$, we cluster these translations into semantic sets $C_i$, each with different meanings. Based on the frequency of each semantic set, we can compute the semantic entropy of $T$:
\begin{equation}
\label{equation1}
    Ent(T) = \text{-}\sum_{i=0}^{n} p_i \log p_i, 
\end{equation}
where $p_i=\frac{|C_i|}{\sum_i^n |C_i|}$ and $n$ is the number of semantic sets decided by the clustering algorithm. Specifically, we adopt the Agglomerative Clustering (AC) algorithm. Its input is derived from the
paraphrase identification model, which assesses the semantic equivalence of all translation pairs. Below, we elaborate on the paraphrase identification model and implementation details.

\textbf{Paraphrase Identification Model.} 
Previous work typically relies on NLI models to judge the equivalence of two sentences. However, based on our observations, these models don't work well in our cases. Different translations often differ only in surface form, with subtle changes that may alter meaning, such as “\textit{Fully enclosed space}” vs. “\textit{All enclosed space}”. LLMs, like ChatGPT, handle these fine-grained distinctions more reliably due to their stronger language understanding. Nevertheless, directly applying LLMs to evaluate large numbers of sentence pairs is prohibitively expensive. To balance accuracy and efficiency, we adopt a distillation-based strategy. In more detail, we sample 200K pairs from all the generated translation sentences and instruct GPT-4.1 to judge their semantic equivalence. These LLM-generated labels are then used to fine-tune an NLI model into a paraphrase identification model. Note that, different from standard NLI setups, we discard entailment relations and frame the task as a binary classification problem with only two labels: equivalent and not equivalent.

\textbf{Implementation Details.} For each source sentence, we first aggregate outputs from all translation models to form a unified translation set. We then apply the fine-tuned paraphrase identification model to assess semantic equivalence for all sentence pairs within this set. Using these pairwise judgments, we employ the AC algorithm to group candidate translations into semantic clusters. Finally, based on these clusters, we compute the semantic entropy of each model's translation set using Equation~\ref{equation1}.

We evaluate models from the Qwen2.5 and Qwen3 families, including both Base and Instruct variants. In addition to semantic entropy (Ent), the average of mean Quality Estimation scores~\footnote{Checkpoint: Unbabel/wmt23-cometkiwi-da-xl} (Avg QE) across all translations in each set, and the average of the best QE score (Max QE) from each translation set are employed as metrics. Moreover, we use the same ambiguous/unambiguous paired sentences employed for ambiguity comparison in Section~\ref{Evaluation Setup}.

\subsection{Evaluation Results and Discussion}
Translation quality and semantic uncertainty are reported in Table~\ref{tab:translation_quality}. None of the evaluated models exhibits a substantial difference in overall translation quality, ensuring that the subsequent analyses are not biased by extremely low-quality translations. Base models tend to achieve higher Max QE scores on the best translation within each set, but exhibit lower Avg QE scores across all translations than their Instruct counterparts. This phenomenon aligns with expectations: Base models produce less controllable outputs and thus higher diversity, which is also reflected in their higher semantic entropy values.

\begin{table}[t]
\centering
\small
\caption{Translation quality and semantic uncertainty on CHA-Gen and CHAmbi. Avg QE denotes the mean quality across all sampled translations; Max QE denotes the best translation per set; Ent denotes semantic entropy (higher indicates greater semantic diversity).}
\label{tab:translation_quality}

\begin{tabular}{l l ccc ccc}
\toprule
 &  & \multicolumn{3}{c}{\textbf{CHA-Gen}} 
     & \multicolumn{3}{c}{\textbf{CHAmbi}} \\
\cmidrule(lr){3-5} \cmidrule(lr){6-8}
\textbf{Family} & \textbf{Model}
 & Avg QE & Max QE & Ent
 & Avg QE & Max QE & Ent \\
\midrule
\multirow{6}{*}{Qwen2.5}
 & 7B-Base      &  0.6531&  0.7698&  1.5278&  0.7281&  0.7993&  0.7584\\
 & 7B-Instruct  &  0.6665&  0.7339&  0.4241&  0.7425&  0.7834&  0.2012\\
 & 14B-Base     &  0.6493&  0.7696&  1.4822&  0.7302&  0.7996&  0.6471\\
 & 14B-Instruct &  0.6792&  0.7338&  0.3439&  0.7451&  0.7808&  0.1567\\
 & 32B-Base     &  0.6573&  0.7682&  1.1653&  0.7341&  0.7994&  0.5579\\
 & 32B-Instruct &  0.6865&  0.7316&  0.2959&  0.7494&  0.7795&  0.1393\\
\midrule
\multirow{6}{*}{Qwen3}
 & 8B-Base      &  0.6576&  0.7672&  1.2532&  0.7358&  0.7990&  0.5785\\
 & 8B-Instruct  &  0.6906&  0.7399&  0.3875&  0.7562&  0.7836&  0.1557\\
 & 14B-Base     &  0.6659&  0.7692&  1.1416&  0.7397&  0.7998&  0.5474\\
 & 14B-Instruct &  0.6946&  0.7352&  0.2926&  0.7603&  0.7819&  0.1198\\
 & 32B-Instruct &  0.6906&  0.7483&  0.4646&  0.7594&  0.7882&  0.1805\\
\bottomrule
\end{tabular}

\end{table}

\textbf{Ambiguous vs Unambiguous Sentences.} We compare the average semantic entropy of translations generated for ambiguous and unambiguous sentences, as reported in Table~\ref{tab:amb_unamb_results}. Across both datasets and all models, translations of ambiguous sentences consistently exhibit higher semantic entropy than those of unambiguous sentences, supporting our hypothesis. However, the magnitude of this gap varies across models. For example, Base models generally yield a larger entropy gap between ambiguous and unambiguous sentences than their instruction-tuned counterparts. This suggests that instruction tuning reduces generation uncertainty, which in turn narrows the entropy gap and induces translation bias.

\begin{table}[htbp]
\small
\centering
\caption{Semantic entropy comparison on translations of ambiguous/unambiguous sets. \textit{Aggregate} denotes the semantic entropy on gathering all models' outputs for each source sentence.}
\label{tab:amb_unamb_results}

\begin{tabular}{llc ccc ccc}
\toprule
 &  &
\multicolumn{3}{c}{\textbf{CHA-Gen}} &
\multicolumn{3}{c}{\textbf{CHAmbi}} \\
\cmidrule(lr){3-5} \cmidrule(lr){6-8}
Family & Model
& Amb. Set & Unamb. Set & Gap
& Amb. Set & Unamb. Set & Gap \\
\midrule

\multirow{6}{*}{Qwen2.5}
 & 7B-Base        & 1.6061 & 1.3071 & 0.2990 & 0.8097 & 0.7329 & 0.0768 \\
 & 14B-Base       & 1.5473 & 1.2986 & 0.2487 & 0.6800 & 0.6308 & 0.0492 \\
 & 32B-Base       & 1.2270 & 0.9914 & 0.2356 & 0.6240 & 0.5252 & 0.0988 \\
\cmidrule(lr){2-8}
 & 7B-Instruct    & 0.4398 & 0.3799 & 0.0599 & 0.2312 & 0.1863 & 0.0449 \\
 & 14B-Instruct   & 0.3614 & 0.2948 & 0.0666 & 0.1672 & 0.1515 & 0.0157 \\
 & 32B-Instruct   & 0.3130 & 0.2479 & 0.0651 & 0.1689 & 0.1246 & 0.0443 \\
\midrule

\multirow{5}{*}{Qwen3}
 & 8B-Base        & 1.3178 & 1.0710 & 0.2468 & 0.6388 & 0.5486 & 0.0902 \\
 & 14B-Base       & 1.1977 & 0.9833 & 0.2144 & 0.6328 & 0.5050 & 0.1278 \\
\cmidrule(lr){2-8}
 & 8B-Instruct    & 0.4144 & 0.3114 & 0.1030 & 0.1798 & 0.1437 & 0.0361 \\
 & 14B-Instruct   & 0.3104 & 0.2423 & 0.0681 & 0.1545 & 0.1026 & 0.0519 \\
 & 32B-Instruct   & 0.5014 & 0.3610 & 0.1404 & 0.2173 & 0.1622 & 0.0511 \\
\midrule

 & \textit{Aggregate}
 & 1.2923 & 1.0134 & 0.2789 & 0.6616 & 0.5304 & 0.1312 \\
\bottomrule
\end{tabular}

\end{table}

\begin{figure}
    \centering
    \includegraphics[width=\linewidth]{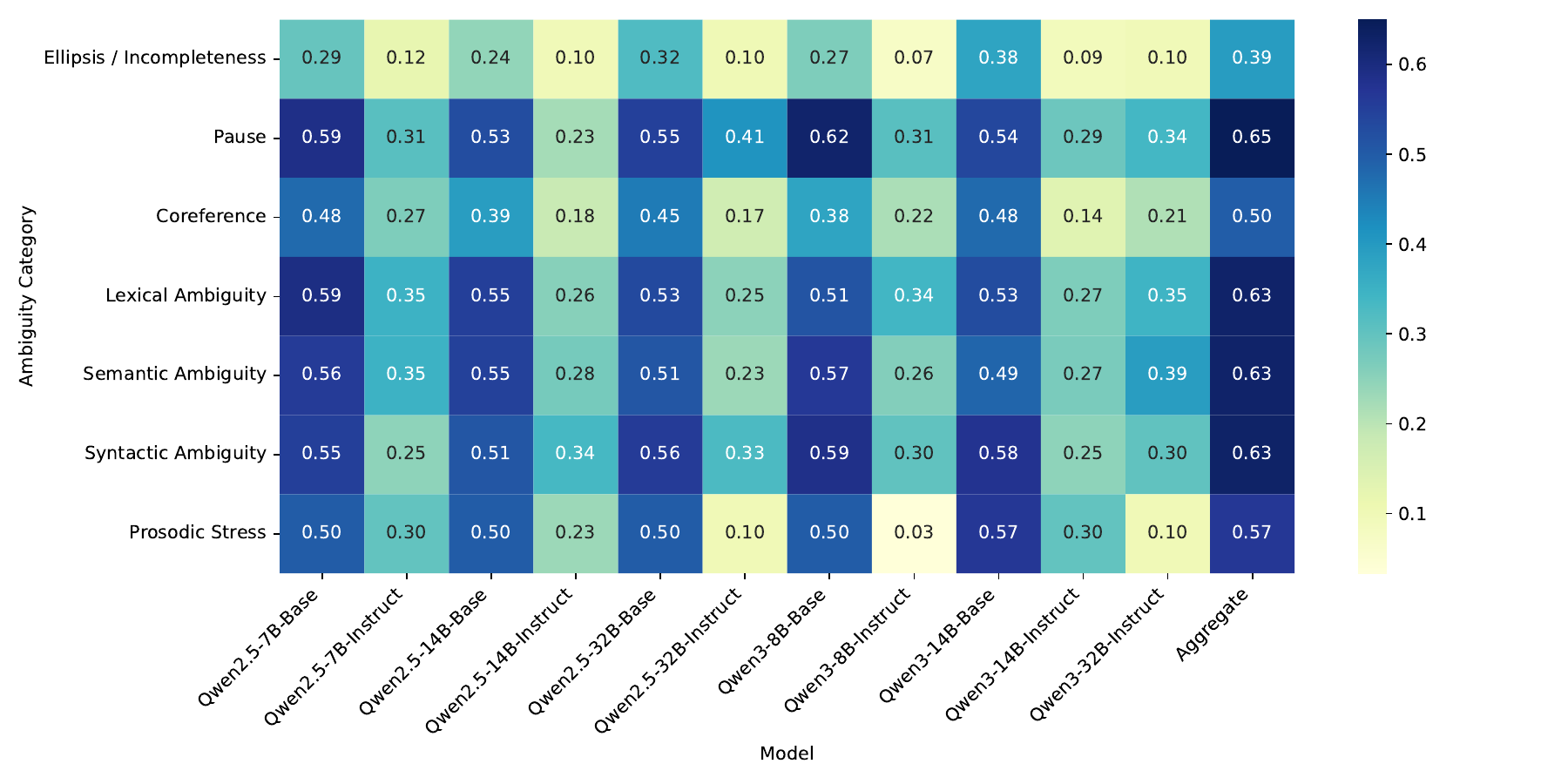}
    \caption{Fraction of CHAmbi sentence pairs where ambiguous sentences exhibit higher semantic entropy in their translation sets than their unambiguous counterparts, broken down by ambiguity types. }
    \label{fig:cat_chambi}
\end{figure}

\textbf{Category-wise Analysis in CHAmbi.} We analyze the fraction of (ambiguous, unambiguous) sentence pairs in CHAmbi for which the ambiguous sentence exhibits higher semantic entropy in its translation set than its unambiguous counterpart, as illustrated in Figure~\ref{fig:cat_chambi}. Among all ambiguity categories described in CHAmbi, incompleteness yields the lowest fraction. This is likely because incompleteness stems from missing information, which induces vagueness rather than supporting multiple well-defined interpretations. The second-lowest fraction is observed for coreference ambiguities. Such ambiguities can often be translated directly into English without resolving pronouns, thereby preserving the original ambiguity and reducing divergence between ambiguous and unambiguous cases. Overall, Base models consistently achieve higher fractions than instruction-tuned variants. This aligns with the trends observed in the previous experiment.

\begin{figure}
    \centering
    \includegraphics[width=0.7\linewidth]{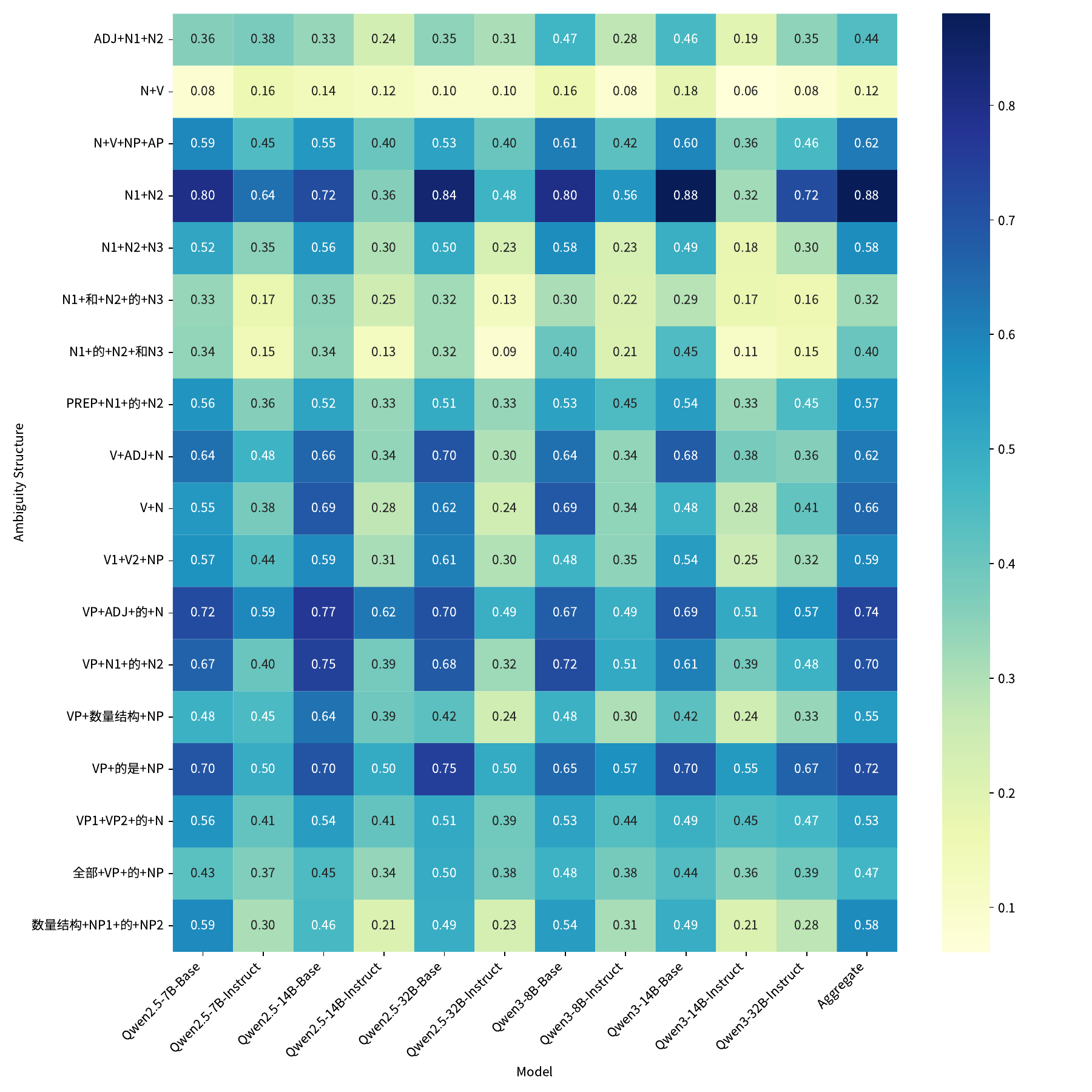}
    \caption{Fraction of CHA-Gen sentence pairs for which ambiguous sentences exhibit higher semantic entropy in their translation sets than their unambiguous counterparts, broken down by ambiguity structures. }
    \label{fig:cat_cn_amb}
\end{figure}

\textbf{Structure-sensitive Analysis in CHA-Gen.} We apply a similar analysis to CHA-Gen, with results shown in Figure~\ref{fig:cat_cn_amb}. Consistent with the CHAmbi results, instruction-tuned models exhibit smaller differences in semantic entropy between ambiguous and unambiguous sentences compared to Base models. Several structural patterns show a high fraction of cases where ambiguous sentences have higher semantic entropy, including “N1+N2,” “VP+ADJ+N,” “VP+N1+的+N2,” and “VP+的是+NP.” This suggests that LLMs are relatively more sensitive to ambiguity arising from these syntactic structures. That is, when inputs involve specific ambiguous syntactic structures, the “confusion” and “hesitation” regarding multiple possible interpretations of the LLM are more strongly reflected as output instability.


\begin{table}[ht]
\centering
\caption{Distribution for different interpretations over the model's generated translation sets of ambiguous Chinese sentences (50 samples per model). For each source sentence, we show two major translation clusters and include one representative translation per cluster. Percentages do not sum to 100\% because translations belonging to other clusters or judged as noisy or low quality are excluded.}
\label{tab:trans_amb_case}
\resizebox{\textwidth}{!}{%
\begin{tabular}{p{3cm} p{6cm} c c c c c c c c}
\toprule
Source sentence & Translations& \multicolumn{2}{c}{Qwen3-8B} & \multicolumn{2}{c}{Qwen3-14B} & \multicolumn{2}{c}{Qwen2.5-7B} & \multicolumn{2}{c}{Qwen2.5-14B} \\
\cmidrule(r){3-4} \cmidrule(r){5-6} \cmidrule(r){7-8} \cmidrule(r){9-10}
& & Base & Instruct & Base & Instruct & Base & Instruct & Base & Instruct \\
\midrule
\multirow{2}{*}{保护主人的猫。} 
& A: The cat that protects its owner. 
& 36\% & 100\% & 4\% & 0\% & 40\% & 100\% & 4\% & 0\% \\
& B: Protect the owner's cat.
& 46\% & 0\% & 86\%& 100\% & 42\% & 0\% & 86\% & 100\% \\
\midrule

\multirow{2}{*}{全部密封的包裹。}
& A: Fully enclosed package.
& 62\% & 100\% & 24\% & 96\% & 18\% & 0\%& 52\% & 100\% \\
& B: All sealed bagages.& 26\% & 0\% & 70\% & 4\% & 60\% & 76\% & 38\% & 0\% \\
\midrule

\multirow{2}{*}{保护的是弱者。}& A: It protects the weak.& 100\%& 100\%& 100\%& 100\%& 92\%& 98\%& 98\%& 100\%\\
& B: It's the weak that protect it.& 0\%& 0\%& 0\%& 0\%& 0\%& 0\%& 0\%& 0\%\\ 
\midrule

\multirow{2}{*}{帮助的是那个老人。}& A: It is the old man who is getting help.& 34\%& 0\%&  22\%& 0\%& 36\%& 22\%& 68\%& 100\%\\
& B: The one who helped was the old man.& 46\%& 100\%& 62\%&  100\%& 12\%& 74\%& 14\%& 0\%\\ 
\midrule

\multirow{2}{*}{帮助的是弱者。}& A: It is the weaks who are helped.& 56\%& 78\%& 62\%& 92\%& 44\%& 70\%& 46\%& 42\%\\
& B: Those who help are the weak.& 14\%& 6\%& 10\%& 8\%& 14\%& 4\%& 4\%& 0\%\\

\bottomrule
\end{tabular}%
}
\end{table}

\subsection{Case Study}

We conduct this case study focusing on two targeted aspects. The first aspect aims to compare the diversity of translations generated by Base models and their Instruct-tuned counterparts, exploring how instruction tuning affects the variability of model outputs. The second aspect focuses on investigating the discrepancies in how different models understand ambiguous linguistic structures, particularly examining the factors that lead models to adopt different interpretive paths for structurally similar or inherently ambiguous sentences. Specifically, we examine several concrete examples to elaborate on these two aspects.

\textbf{Base vs Instruct.} The first two examples in Table~\ref{tab:trans_amb_case} illustrate differences in interpretation distributions across model variants. In these cases, Base models exhibit greater translation uncertainty and diversity than their Instruct counterparts. For instance, Qwen2.5-7B-Base produces a nearly balanced distribution between Translation A and Translation B for “保护主人的猫。”, while Qwen2.5-14B-Base shows a similar balance for “全部密封的包裹。”. After instruction tuning, however, most models become strongly biased toward a single interpretation. These observations suggest that the uncertainty inherent in ambiguous sentences is better preserved by Base models pretrained on large-scale Internet corpora, whereas instruction tuning tends to suppress such uncertainty and promote deterministic interpretations.

\textbf{Dominant Interpretation Bias.} The last three example sentences in Table~\ref{tab:trans_amb_case} share the structure “VP + 的是 + NP”, such as “保护(protect)的是(is)弱者(the weak)”, “帮助(help)的是(is)那个老人(that old man)”, and “帮助(help)的是(is)弱者(the weak)”. Such structures may induce ambiguity when the agent and patient roles of the action are unspecified.

Despite their high syntactic similarity, models exhibit distinct interpretation behaviors across these three sentences. Specifically, for sentences involving “弱者 (the weak)”, “保护的是弱者” and “帮助的是弱者”, models show a strong tendency to parse “弱者” as the patient of the action (i.e., the recipient of protection or help). This preference stems from the cognitive habit in daily language where “弱者” is often the action recipient. “Protecting the weak” and “helping the weak” are far more frequent in real-world corpora than alternative interpretations (e.g., “the weak protecting others” or “the weak helping others”).

This leads to two key findings. First, even with identical syntactic structures, the dominant status of specific interpretations in common usage reduces a sentence’s perceived ambiguity. From a formal semantic perspective, however, such sentences still retain theoretically multiple interpretive possibilities in specific contexts. Second, the observed model bias indicates that LLMs tend to choose frequent or fixed interpretive paths rather than fully capturing and expressing the inherent ambiguous space of sentences.

Furthermore, this behavioral pattern explains why not all theoretically ambiguous sentences in our dataset (even in Base models) exhibit obvious translation uncertainty. In particular, sentences differ in degree of interpretive ambiguity, which stems from the frequency distribution of their possible interpretations in natural language. This variation in interpretive ambiguity directly modulates models' ability to recognize and characterize ambiguity. Such over-reliance on dominant interpretations undermines the robustness of LLMs in handling truly ambiguous sentences.

This finding aligns with previous work~\citep{itzhak-etal-2024-instructed} showing that instruction tuning can introduce systematic cognitive biases in model behavior. While instruction tuning substantially improves model usability and alignment with user expectations, it may unintentionally amplify preferences for dominant or high-frequency interpretations, thereby suppressing alternative plausible readings. In real-world cross-lingual communication, contexts are considerably more complex. Therefore, ambiguities are often implicit and considerably harder to detect than those in our controlled experimental settings. Consequently, such biases may result in misunderstandings or misrepresentations.

On the other hand, our observation that Base models exhibit fewer systematic biases and produce more diverse outputs suggests a possible direction for mitigation. Beyond instruction tuning alone, incorporating the diversity of Base model generations could serve as an auxiliary signal to retain interpretive multiplicity.

\section{Conclusion}
\label{sec6}

In this paper, we present CHA-Gen, a novel corpus designed to enrich existing resources for Chinese ambiguity research. Built on PA theory, it is systematically constructed via a pipeline integrating automated data curation and rigorous manual verification. This design effectively achieves a balance between scalability and annotation accuracy. The corpus comprises 2,414 ambiguous and 3,298 unambiguous sentences, covering 18 distinct ambiguous structures to support comprehensive LLM ambiguity evaluation.

Evaluations are conducted through complementary approaches: direct querying and machine translation. The former assesses LLMs’ ability to detect and discriminate ambiguity, while the latter probes their interpretive uncertainty. Overall results demonstrate that Chinese ambiguity remains highly challenging for current LLMs. In more detail, 1) direct querying evaluation shows that better general LLM capabilities (larger scale or higher versions) do not guarantee improved ambiguity detection. While CoT prompting balances predictions and reduces bias, rationales reveal three failure patterns: ambiguity blindness, misattribution, and premature resolution. This study indicates a divergence from human judgment and provides insights for future improvement. 2) Machine translation evaluation demonstrates that specific ambiguity categories (e.g., incompleteness, coreference) and certain syntactic structures (e.g., N1+N2, VP+ADJ+N) induce higher semantic entropy. Instruction tuning tends to suppress output uncertainty, and a dominant interpretation bias persists across models. This further highlights their limited ability to capture the full linguistic ambiguity and the concerns of misunderstanding in cross-lingual communication. 

In conclusion, this study reveals the limitations of current LLMs in handling Chinese ambiguity, emphasizing the necessity of improving evaluation strategies and conducting targeted model optimizations. We hope that the proposed dataset and estimation methods will support and accelerate progress in this research area.

\section{Limitations}

Our constructed corpus currently focuses on syntactic ambiguity, which allows for more controlled and consistent data generation. While this enables focused analysis, a more comprehensive evaluation would benefit from incorporating additional types of ambiguity, such as polysemy, word segmentation, and other linguistic phenomena. Another potential limitation concerns sentence length. Our corpus primarily consists of relatively short utterances and phrases. Although such examples support analysis across different levels of linguistic complexity, real-world scenarios frequently involve longer and more syntactically complex sentences. In future work, we plan to explore (semi-)automatic strategies to generate and expand the corpus with more diverse and complex ambiguous instances toward a more holistic evaluation.

\FloatBarrier

\appendix

\section{Details of Datasets}
\label{dataset_examples}

\subsection{Structure/syntax-based Prompts for Corpus Construction}
\label{prompt_template}
 Structure-based prompts and syntax-based prompts for CHA-Gen corpus construction described in Section~\ref{Corpus Construction} are given in Table~\ref{tab:dataset-syntax-based} and Table~\ref{tab:structure-based_prompt_template}, respectively.

\begin{table}[!p]
\caption{Examples of syntax-based prompts for CHA-Gen corpus construction.}
\label{tab:dataset-syntax-based}

\resizebox{\textwidth}{!}{%
\begin{tabular}{p{3cm} p{12cm}}
\toprule
\textbf{Potential Ambiguous Structure} & \textbf{Prompt} \\
\midrule
VP1+VP2+的+N & 
生成20句VP+的+N结构的中文句子，并在每个句子前分别加一个动词，使句子符合逻辑，
例如：看打球的学生。
词汇尽量丰富，生成的句子尽量不重复，生成的句子间用"。"结尾且进行编号如1.

(Generate 20 sentences with the structure "VP + 的 + N," where a verb is added before each sentence to make it logically complete. An example sentence is: "看打球的学生(Students who watch ball games\textbackslash Watch students playing ball games)". Each sentence is numbered and ends with a period.)\\
\midrule
N+V+NP+AP &
生成20个的N1+V+N2的短语，并在后面加上一个简单形容词来又能形容N2和形容短语，即形成N1+V+N2+ADJ的句子，使句子符合逻辑，
例如：张三笑李四很笨。 
词汇尽量丰富，生成的句子尽量不重复，生成的句子间用"。"结尾且进行编号如1.

(Generate 20 sentences with the structure "N1 + V + N2 + ADJ," where an adjective is added to describe both N2 and the entire phrase. An example sentence is: "张三笑李四很笨(Zhang San laughs at Li Si for being very foolish\textbackslash It is foolish that Zhang San laughs at Li Si)". Each sentence is numbered and ends with a period.)\\
\bottomrule
\end{tabular}
}
\end{table}

\subsection{Validation Prompt for Corpus Construction}
\label{validation_prompt}


\begin{table}[htbp]
\caption{Examples of structure-based prompts for CHA-Gen corpus construction.}\label{tab:structure-based_prompt_template}
\resizebox{\textwidth}{!}{%
\begin{tabular}{p{3cm} p{12cm}}
\toprule
\textbf{Potential Ambiguous Structure} &  \textbf{Prompt}\\
\midrule
N1+N2+N3 &
生成10句N1+N2+N3结构的句子，
要求N1一定是名词，N3既能被N1修饰也能被N2修饰，整句话只有3个名词，
例子：进口飞机引擎。
生成的句子间用"。"结尾且进行编号如1.

(Generate 10 sentences with the structure N1 + N2 + N3, where N1 is a noun, N3 can be modified by either N1 or N2, and each sentence contains exactly three nouns. An example sentence is: "进口飞机引擎(Imported aircraft engine\textbackslash Engine of imported aircraft)". Each generated sentence is numbered and ends with a period.)\\
\midrule
VP+的是+NP & 
生成10句VP+的+是+NP结构的句子，
要求VP既能作主动，也能作被动，NP是有生命的物体，
例子：反对的是少数人。
生成的句子间用"。"结尾且进行编号如1.

(Generate 10 sentences with the structure "VP + 的 + 是 + NP," where the VP (verb phrase) can function both actively and passively, and the NP (noun phrase) must be a living entity. An example sentence is: "反对的是少数人(Those who oppose are a minority\textbackslash A minority is being opposed)". Each generated sentence is numbered and ends with a period.)\\ 
\bottomrule
\end{tabular}
}
\end{table}

Table~\ref{tab:dataset-validation-prompt} gives the validation prompt and schema used in CHA-Gen corpus construction. 


\begin{table}[htbp]
    \caption{Prompt Template for Ambiguity Validation}  
    \label{tab:dataset-validation-prompt}
    \resizebox{\textwidth}{!}{%
    \begin{tabular}{p{18cm}}  
        \toprule
    请你根据输入的输入的歧义结构“ambType”和歧义理由“ambReason”判断“source”这个句子是否具有歧义，你的输出需要包含以下信息：ambiguity：判断句子是否存在歧义，如果存在歧义，将其设置为true，否则设置为false。reasons：解释句子产生歧义的原因。example：根据歧义类型和理由，生成多个可能的上下文，展示句子的不同含义。每个上下文必须完整包含原句。输出时仅保留括号{}内的内容.\\
    (Please judge whether the sentence "source" is ambiguous according to the Potential Ambiguous Structure "ambType" and linguistic explanation "ambReason" of the input. Your output needs to include the following information:ambiguity: judge whether the sentence is ambiguous. If there is ambiguity, set it to true, otherwise set it to false. Reasons: explain the reasons why sentences are ambiguous. Example: generate multiplepossible contexts to show the different meanings of sentences according to the types and reasons of ambiguity. Each context must contain the original sentence completely. Only the contents in brackets \{\} are retained when outputting)\\
    \\
    输出应该以符合JSON模式的JSON实例进行格式化输出：\\
    (The output should be formatted as a JSON 
    instance that conforms to the JSON schema 
    below:)\\
    \{"properties" : \{"foo": \{"title": "Foo", "description": "list of integer", "type": "array", "items": \{"type": "integer"\}\}\}, "required": ["foo"]\}. The object \{"foo": [0, 1]\} is a well-formatted instance of the schema. The object \{"properties": \{"foo": [0, 1]\}\} is not well-formatted.\\ 
    \\
    这是我们任务输出的格式：\\ 
    (This is the schema of our task output:)\\
    \{"properties": \{"ambiguity": \{"title": "Ambiguity", "description": "bool","type": "bool"\}, "reasons": \{"title": "Reasons", "description": "text", "type": "string"\}, "examples": \{"title": "Examples", "description": "array of ambiguity examples", "type": "array", "items": \{"type": "string"\}\}\}, "required": ["ambiguity", "reasons", "examples"]\}\\
    \\
    接下来请回答正式输入：\\
    (Next, please answer the formal input:)\\
    \bottomrule
    \end{tabular}
    }
\end{table}

\FloatBarrier 

\subsection{CHA-Gen Corpus}
\label{CHA-Gen_ambiguity_structure}
Table~\ref{tab:ambi-structure} presents the Potential Ambiguous Structures and the corresponding examples of CHA-Gen. The notations "VP, NP, N, Q, ADJ, V, PREP" stand for "Verb Phrase, Noun Phrase, Noun, Quantifier, Adjective, Verb, Preposition", respectively.

\begingroup
\small 
\begin{longtable}{m{3.5cm} m{4cm} m{7.5cm}}
\caption{\small Potential Ambiguous Structures and examples of the CHA-Gen corpus.} \label{tab:ambi-structure} \\
\toprule
\textbf{Potential Ambiguous Structure} & \textbf{Example} & \textbf{Linguistic explanation} \\
\midrule
\endfirsthead

\multicolumn{3}{c}{\tablename\ \thetable{} (Continued)} \\
\toprule
\textbf{Potential Ambiguous Structure} & \textbf{Example} & \textbf{Linguistic explanation} \\
\midrule
\endhead

\midrule
\multicolumn{3}{r}{\textit{Continued on next page...}} \\
\endfoot

\bottomrule
\endlastfoot

VP+的是+NP &
反对的是少数人 
(Those who oppose are a minority \textbackslash A minority is being opposed) & 
句子可以理解为少数人作为动作主体进行反对，也可理解为被反对的对象是少数人 
(The sentence can be understood as a minority acting as the subject of opposition, or as the object of opposition being the minority) \\
\midrule

N1+N2+N3 & 
进口飞机引擎 
(Imported aircraft engine \textbackslash\ Engine of imported aircraft) & 
句子可以理解为进口的飞机引擎，也可以理解为进口飞机的引擎
(The sentence can be understood as an imported aircraft engine, or as an engine of the imported aircraft) \\
\midrule

VP+Q+NP & 
分析了十分钟数据 
(Data were analyzed for ten minutes \textbackslash\ A ten-minute data was analyzed) & 
句子中的“十分钟”既可以理解为分析动作持续的时间（分析用了十分钟），也可以理解为被分析的数据内容（十分钟长度的数据） 
(The phrase 'ten minutes' in the sentence can be understood as both the duration of the analysis action (which took ten minutes) and the content of the analyzed data (which is ten minutes long)) \\
\midrule

N1+N2 & 
牛奶面包 
(Milk-flavored bread \textbackslash\ Milk and bread) & 
句子可以理解为并列关系的两种食物（牛奶和面包），也可理解为用牛奶制作的面包（偏正结构）
(The Sentence can be understood as two parallel foods (milk and bread), or as bread made from milk (with a skewed structure)) \\
\midrule

Q+NP1+的+NP2 & 
三个学校校长 
(principal from three schools \textbackslash\ Three principals from a school) & 
句子可指3所学校的校长（每校1人）或3名担任校长职务的人（可能属于同一学校）
(The sentence can refer to the principals of three schools (one person per school) or three individuals holding the position of principal (who may belong to the same school)) \\
\midrule

ADJ+N1+N2 & 
小型宠物商店 
(Small pet store \textbackslash\ Store for small pets) & 
句子可以理解为小型的（宠物商店），也可以理解为（小型宠物）的商店
(The sentence can be understood as a small (pet store) or a store for (small pets)) \\
\midrule

VP1+VP2+的+N & 
看打球的学生 
(Students who watch ball games \textbackslash\ Watch students playing ball games) & 
句子既可以理解为学生在看打球（的比赛），也可以理解为他人看这些打球的学生。
(The sentence can be understood as students watching basketball games, or as watching students playing games.) \\
\midrule

V1+V2+NP & 
评估改善工作流程 
(Evaluate and improve workflow \textbackslash\ Evaluate improved workflow) & 
短语结构存在双重解析可能：1. “评估改善后的工作流程”（评估对象为已被改善的流程） 2. “评估并改善工作流程”（并列动作）
(There is a possibility of dual parsing in phrase structures: 1 Evaluate the improved workflow (evaluating the improved process) 2 Evaluate and improve workflow (parallel action)) \\
\midrule

N1+和+N2+的+N3 & 
桌子和椅子的腿 
(Legs of both tables and chairs \textbackslash\ Chairs and legs of tables) & 
句子可以理解为桌子的腿和椅子腿，也可以理解为桌子椅子的共有腿
(The sentence can be understood as a leg of a table and a leg of a chair, or as the common leg of a table and a chair) \\
\midrule

V+N & 
翻译文件 
(Translate documents \textbackslash\ Translated documents) & 
句子既可表示翻译的动作行为，也可指用于翻译的特定文件
(The sentence can represent both the action behavior of translation and specific translated documents) \\
\midrule

N1+的+N2+和+N3 & 
手机的屏幕和外壳 
(The screen of a mobile phone and a casing \textbackslash\ The screen and casing of a mobile phone) & 
“手机”可以同时修饰“屏幕”和“外壳”(联合修饰)，也可以仅修饰“屏幕”而让“外壳”成为独立并列项(单边修饰)。
(The 'phone' can modify both the 'screen' and the 'casing' (joint modification), or it can modify only the 'screen' and make the 'casing' a separate item (unilateral modification)) \\
\midrule 

N+V & 
软件开发 
(Software development \textbackslash\ Develop a software) & 
句子既可以作为整体名词概念指代软件工程领域，也可以理解为名词+动词结构表示对软件进行开发的动作过程（对软件进行开发）
(The sentence can be used as a holistic noun concept to refer to the field of software engineering, or it can be understood as a noun verb structure to represent the action process of developing software (developing software)) \\
\midrule

PREP+N1+的+N2 & 
关于老师的意见 
(Opinions about the teacher \textbackslash\ About teacher's opinions) & 
可以理解为关于（老师所做的意见），也可以理解为这个意见是关于老师的
(The sentence can be understood as about (the teacher's opinion), or it can be understood as an opinion about the teacher) \\
\midrule

VP+ADJ+的+N & 
照顾周到的妈妈 
(Thoughtful mother \textbackslash\ Taking care of thoughtful mother) & 
句子可以表示照顾（周到的妈妈），“照顾”修饰“周到的妈妈”形成动宾结构; 也可以表示（照顾周到的）妈妈，理解为妈妈是照顾周到的 
(The sentence can express care (thoughtful mother), and "care" modifies "thoughtful mother" to form an object verb structure; It can also mean a thoughtful mother, understood as a mother who is thoughtful) \\
\midrule

VP+N1+的+N2 & 
拥抱父亲的儿子 
(The son who embrace his father \textbackslash\ Embracing father's son) & 
既可以理解为（拥抱父亲）的儿子，也可以理解为拥抱（父亲的儿子）
(The sentence can be understood as the son who embrace his father, or as embracing the son of the father) \\
\midrule

V+ADJ+N & 
晒干毛巾 
(Sun dried towels \textbackslash\ Sunbathing dry towels) & 
句子可理解为：“晒/干毛巾”（将干毛巾拿去晾晒）或“晒干/毛巾”（通过晾晒使毛巾变干）
(The sentence can be understood as: 'Drying/drying towels' (taking dry towels to air dry) or' sun drying/drying towels' (drying towels by air drying)) \\
\midrule

N+V+NP+AP & 
同学评价老师很严格 
(The classmate evaluated the teacher for being very strict \textbackslash\ Classmates evaluate teachers as very strict) & 
句子可以理解为同学对“老师很严格”进行评价或者同学评价老师的行为本身很严格
(The sentence can be understood as a classmate's evaluation of 'the teacher is very strict' or a classmate's evaluation of the teacher's behavior as being very strict) \\
\midrule

全部 \textbackslash\ 部分+VP+的+NP & 
全部打开的窗户 
(A completely opened window \textbackslash\ All of the opened windows) & 
句子可以理解为所有被打开了的窗户，也可以理解为窗户被完全打开
(The sentence can be understood as all the windows that have been opened, or as windows that have been fully opened) \\

\end{longtable}

\endgroup

\section{Prompts for Translation Sampling}
\label{apd:sample_trans}
The prompts used for sampling English translations are presented in Table~\ref{tab:translation_prompts}. To better control the model outputs, we enclose the expected translations in double quotation marks. For Instruct models, after applying the corresponding chat templates, we additionally prepend “\texttt{English:"}” to further constrain the generated outputs.

\begin{table}[h]
\centering
\caption{Prompt templates used for sampling English translations.}
\label{tab:translation_prompts}

\begin{subtable}{\linewidth}
\caption{Template for Base models}
\label{tab:translation_prompt_base}

\begin{promptbox}{}
\ttfamily
Translate this from Chinese into English: \\
Chinese: "\{src\}" \\
English: "
\end{promptbox}

\end{subtable}

\vspace{0.5em}

\begin{subtable}{\linewidth}
\caption{Template for Instruct models with the corresponding chat template (Qwen3 as an example)}
\label{tab:translation_prompt_instruct}

\begin{promptbox}{}
\ttfamily
<|im\_start|>user\\
Translate this from Chinese into English:\\
Chinese: "\{src\}"<|im\_end|>\\
<|im\_start|>assistant\\
<think>\\
\\
</think>\\
\\
English: "
\end{promptbox}

\end{subtable}

\end{table}

\section{Declaration of generative AI}
During the preparation of this work, the authors used ChatGPT and DeepSeek as writing-support tools to improve clarity and refine the language, and used Gemini Nano Banana to enhance sketches and generate small illustrations. All content was carefully reviewed and revised by the authors, who take full responsibility for the final version.


\clearpage
\bibliographystyle{cas-model2-names}

\bibliography{custom,anthology_small}



\end{document}